\documentclass[conference]{IEEEtran}
\usepackage{comment}

\IEEEoverridecommandlockouts
\usepackage{cite}
\usepackage{url}
\usepackage{enumitem}
\usepackage{amsmath,amssymb,amsfonts}
\usepackage[linesnumbered,ruled,vlined]{algorithm2e}
\usepackage{flushend}

\usepackage{graphicx}
\usepackage{textcomp}
\usepackage{xcolor}
\usepackage{pgfplots}
\usepackage{pgfplotstable}
\pgfplotsset{compat=newest}
\usepgfplotslibrary{groupplots} % Load the required library
\usepackage{tikz}
\usepackage{subcaption}
\usepackage{float}
\usepackage{booktabs}  
\usepackage{array}  
\usepackage{makecell}  % For easy multi-line alignment

\newcommand{\mahmoud}[1]{\textcolor{blue}{Mahmoud says: {#1}}}

\newcommand{\gaspard}[1]{\textcolor{blue}{Gaspard says: {#1}}}

\pgfplotsset{
    discard if/.style 2 args={
        x filter/.code={
            \edef\tempa{\thisrow{#1}}
            \edef\tempb{#2}
            \ifx\tempa\tempb
                
            \fi
        }
    },
    discard if not/.style 2 args={
        x filter/.code={
            \edef\tempa{\thisrow{#1}}
            \edef\tempb{#2}
            \ifx\tempa\tempb
            \else
                
            \fi
        }
    }
}

\def\BibTeX{{\rm B\kern-.05em{\sc i\kern-.025em b}\kern-.08em
    T\kern-.1667em\lower.7ex\hbox{E}\kern-.125emX}}
\begin{document}

\title{Using LLMs for Analyzing AIS Data\\
\thanks{Identify applicable funding agency here. If none, delete this.}
}

\author{\IEEEauthorblockN{1\textsuperscript{st} Gaspard Merten}
\IEEEauthorblockA{\textit{Data Science and Engineering Lab} \\
\textit{Université libre de Bruxelles}\\
Brussels, Belgium \\
gaspard.merten@ulb.be}
\and
\IEEEauthorblockN{2\textsuperscript{nd} Gilles Dejaegere}
\IEEEauthorblockA{\textit{Data Science and Engineering Lab} \\
\textit{Université libre de Bruxelles}\\
Brussels, Belgium \\
gilles.dejaegere@ulb.be}
\and
\IEEEauthorblockN{3\textsuperscript{rd} Mahmoud Sakr}
\IEEEauthorblockA{\textit{Data Science and Engineering Lab} \\
\textit{Université libre de Bruxelles}\\
Brussels, Belgium \\
mahmoud.sakr@ulb.be}
}

\maketitle
\begin{abstract}
Recent research in Large Language Models (LLMs), has had a profound impact across various fields, including mobility data science. This paper explores the and experiment with different approaches to using LLMs for analyzing AIS data. We propose a set of carefully designed queries to assess the reasoning capabilities of LLMs in this kind of tasks. Further, we experiment with four different methods: (1) using LLMs as a natural language interface to a spatial database, (2) reasoning on raw data, (3) reasoning on compressed trajectories, and (4) reasoning on semantic trajectories. We investigate the strengths and weaknesses for the four methods, and discuss the findings. The goal is to provide valuable insights for both researchers and practitioners on selecting the most appropriate LLM-based method depending on their specific data analysis objectives.
\end{abstract}


\section{Introduction and Motivation}
The significant development in artificial machine learning has also opened the way to new approaches to solve real-world geospatial problems. %\cite{vopham2018emerging, mokbel2023towards, graser2024mobilitydl}.
In particular, Large Language Models (LLMs) have emerged as powerful tools for understanding and generating human-like text. 
These models have demonstrated remarkable abilities in natural language processing tasks, from answering complex queries to summarizing and interpreting information in various domains.% \cite{goyal2024healai, sallam2023utility}.

This exponential increase of LLMs usage can also be witnessed in the domain of Geographic Information Systems (GIS) in recent years.
We focus on a particular subset of GIS managing mobility (or spatio-temporal) data.
Many researchers have proposed tools that interact with GIS tools \cite{li2023autonomous, zhang2024geogpt}.
The main ideas behind these tools consist in using the LLMs to build a pipeline of operations to be executed by the ad-hoc GIS tools (PostGIS databases, PyQGIS scripts, etc.) to provide the results of the processing.
In 2023, Manvi et al. proposed another approach. 
They investigated the capacities of LLMs to answer geospatial queries and the improvement of results by enriching prompts with data related to the query and fetched from OpenStreetMap.
Finally, multiple custom-fine-tuned LLMs have been developed and specialized to answer geoscience questions \cite{deng2024k2, lin2023geogalactica}.
Shortcomings such as biases or inaccuracies of LLM have also been pointed out for general usage and also in the geoscience field \cite{janowicz2023philosophical}.
% philosophical, highlights biases and inaccuracy of llms. Could be used to justify our approach where we do not show a "better solution" but instead compare different approaches and try to hint what works better with which approach and to which extend.

In another perspective, the widespread adoption of AIS has resulted in the accumulation of vast amounts of ship trajectory data.
While AIS data has initially been developed for collision avoidance, it is nowadays widely used to exchange various navigational information between ships, terrestrial stations, and satellites. 
Analyzing this data is crucial for understanding maritime activities and supports a wide range of real-world applications, including estimating greenhouse gas emissions, constructing maritime transport networks, detecting (illegal) fishing activities, and predicting future vessel movements.  %\cite{rindone2024ais, wu2022semantic, hexeberg2017ais}. 

Effective analysis of AIS data requires integration with other sources. For instance, in \cite{wu23CO2}, AIS data was fused with the engine information of ships to estimate CO2 emission. LLMs are pretrained with vast amounts of world information. They encode in their parameters a comprehensive view of knowledge about all domains of our lives, which positions them as extensive knowledge bases. This paper aims to investigate the use of LLMs in the analysis of AIS data.

Using Large Language Models (LLMs) to query and analyze data has become a prominent application, exploring their ability to understand and execute data analysis tasks. Despite the scarcity of this literature, due to the recent introduction of LLMs, we could distinguish different approaches as follows:

\noindent\textbf{(1) Natural Language Interfaces to Databases (NLIDB):} LLMs are used to translate natural language queries into SQL statements or Python scripts. This application is especially useful in making data accessible to users without formal training in database querying and programming. This approach is commonly known as NL2SQL~\cite{NL2SQL24}, and it explores methods to improve the translation of natural language to SQL through deep learning models. This approach has been proposed for geospatial and mobility data, e.g., in~\cite{10.1145/3589132.3625597}, suggesting that integrating LLMs into spatial data management could lead to an innovative database system that learns from both structured and unstructured spatial data. Such a system would provide effortless access to spatial information, benefiting not just individual users but also businesses and government policymakers. 
    
\noindent\textbf{(2) Zero-shot data analysis (ZSA):} LLMs like GPT-3 enable these models to perform data analysis tasks they haven't been explicitly trained for, without additional fine-tuning. This capability stems from their extensive pre-training on diverse internet texts, which equips them with a broad understanding of language and various knowledge domains. In zero-shot data analysis, LLMs utilize this pre-trained knowledge to interpret natural language queries about datasets, allowing them to generate insights, summarize trends, and even hypothesize based on the data presented. 
    
\noindent\textbf{(3) Fine-tuning a pre-trained model:} An LLM can be extended by continuing its training on a smaller, task-specific dataset. This method is essential for adapting general models to perform effectively on specific tasks by refining their parameters to better meet those tasks' unique requirements. Unlike zero-shot learning, which requires no additional task-specific data and relies on general pre-trained knowledge, fine-tuning demands a labeled dataset for the specific task, leading to higher accuracy and performance. While zero-shot learning allows for immediate deployment across various tasks, making it quicker and more cost-effective for less critical scenarios, fine-tuning offers greater customization and optimization, which is vital for more complex or crucial tasks.

\noindent\textbf{(4) Between zero-shot learning and fine-tuning, there are intermediate approaches like few-shot learning and transfer learning.} Few-shot learning operates between the extremes, utilizing very few labeled examples to guide the model—more than zero but significantly fewer than in typical supervised learning. Transfer learning, while similar to fine-tuning, generally involves adjusting a pre-trained model to a new but related problem or domain and can sometimes include the initial adaptation of a general model to a specific field before any detailed fine-tuning.

In this paper, we investigate the first two approaches; namely: Natural Language Interfaces to Databases (NLIDB) and zero-shot data analysis (ZSA), particularly in the context of analyzing Automatic Identification System (AIS) data. In other words, we investigate using an out-of-the-box LLM, rather than extending it. Recently, Money et al. \cite{mooney2023towards} have presented a study of the capacities of ChatGPT to answer generic geospatial questions. Instead, we focus on the abilities of an LLM to analyse a provided dataset. We also investigate the potential improvement of the LLM reasoning capacities by transforming raw data into a semantic description, closer to natural language (similarly as what has been done by Liang et al. in 2023) or by compressing the trajectories. 

One challenge in the field of AIS data analysis is the absence of a standard benchmark. Further, previous research often relies on queries that are direct mapping into corresponding SQL commands, such as instructing an LLM to locate restaurants within a specific area (range query) or to identify the nearest-neighbor features near a given location, e.g., \cite{qi2023maasdb}. We argue that such queries do not fully leverage the intrinsic knowledge that LLMs acquire from the extensive datasets used during their pre-training. To address this gap, our paper proposes a set of queries organized into a taxonomy specifically designed to cover a wider spectrum of AIS data analysis tasks. 

This work does not pretend to propose a superior solution to solve AIS data analytics problems. Instead, we investigate and compare different approaches to using LLMs in this task. We aim both to illustrate the capacities and shortcomings of the different approaches to using LLMs and to help practitioners select the appropriate approach for their problems. Points of attention when developing LLM based GIS applications are also highlighted. The contributions of this work are as follows:

\begin{itemize}[leftmargin=*]
    \item \textbf{Queries:} We designed 27 analysis questions, in four classes for testing the LLM reasoning capabilities in AIS data. These queries are not limited to direct translations of GIS functions. Further, we established ground truth for these queries in various dataset sizes. For the majority of queries, an automated script (also in the public repository~\footnote{https://github.com/GaspardMerten/AISLLMPaper}) facilitated the computation of ground truth, enabling efficient application to additional datasets. However, in cases where determining ground truth was particularly challenging, domain experts conducted manual verification. Consequently, the ground truth in these instances is not transferable.
    \item \textbf{Experimental evaluation}: of four alternative methods for using LLMs to analyze AIS data. We investigate four LLM-based methods to answer queries related to AIS data, and present the result in this paper.
\end{itemize}

\section{Related works}

Related work can be grouped into three categories.
First, trajectory representation learning focuses on creating embeddings that capture the essential spatiotemporal characteristics of trajectories. \cite{fu2020trembr} utilizes Recurrent Neural Networks (RNNs) and incorporates road network constraints to improve embedding quality. \cite{8509283} presents a deep learning approach designed to be robust to low-quality trajectory data, such as sparse or noisy samples. \cite{7447767} incorporates a wide range of contextual factors (user, trajectory, location, and time) into its embedding model. \cite{10.1145/3557915.3560972} presents a vision for a unified, BERT-like system capable of handling a variety of trajectory analysis tasks. While these works are foundational for understanding and representing trajectories, they do not employ LLMs for direct query answering based on spatiotemporal data.

Second, traditional Natural Language Interfaces to Databases (NLIDBs) aim to allow users to query databases using natural language instead of formal query languages. \cite{affolter2019comparative} provides a categorization of different NLIDB approaches, highlighting the trade-offs between their expressiveness and complexity. \cite{katsogiannis2023survey} focuses specifically on deep learning-based text-to-SQL systems, which translate natural language into SQL queries. More recently, \cite{hong2024next} reviewed the usage of LLM, instead of more traditional NLP approaches, to generate SQL from text queries.

Finally, and most relevant to our work, is the emerging field of LLMs for spatiotemporal data understanding and generation. \cite{liu2024can} introduces STG-LLM, which uses a specialized tokenizer and adapter to enable LLMs to perform spatiotemporal forecasting. \cite{jin2023large} provides a comprehensive survey of large models applied to time series and spatiotemporal data, offering a broad overview of the field. \cite{lan2024traj} explores the use of LLMs for trajectory prediction, leveraging the LLM's capabilities for scene understanding and incorporating lane-aware probabilistic learning. \cite{zhang2024geogpt} presents an LLM-based Geographic Information System (GIS) but still relies on traditional database querying involving spatial databases such as PostGIS. \cite{jiawei2024large} introduces an LLM agent framework for generating personal mobility patterns, focusing on simulation rather than query answering. \cite{10.1145/3637528.3671709} introduces a benchmark to investigate the capabilities of LLMs for Dynamic graphs, hence its spatio-temporal reasoning abilities. \cite{qi2023maasdb} is a vision paper proposing the development of LLM-based spatial database systems. While these works demonstrate the growing potential of LLMs in the spatiotemporal domain, they often concentrate on specific tasks like forecasting or prediction, or they still rely on interactions with traditional GIS tools.

Existing research, therefore, primarily focuses on either representing spatiotemporal data for specific tasks or translating natural language into formal database queries.  Even recent work utilizing LLMs often relies on structured interactions with external GIS tools or databases, or addresses tasks other than direct question answering. This highlights a gap in the literature: a systematic investigation into the capabilities of LLMs to directly reason about and answer natural language queries based on raw spatiotemporal data, without the intermediary of a formal query language or a traditional database system. Our work addresses this gap by exploring this direct LLM reasoning approach and comparing its performance against established NLIDB methods.

\section{Taxonomy of maritime queries}\label{sec:queries}
%This experimental study focus on the evaluation of different LLM based frameworks according to their answers to queries related to the maritime domain.
%More precisely, we focus on how these frameworks can both retrieve data, compute estimates or perform more elaborated reasoning on enriched AIS data \gilles{Gaspard: insert citation}.

We use multiple data sources that are commonly available for Maritime data scientists: (1) Raw AIS data in the form of a comma separated table~\footnote{AIS data for the 20/11/2024 from the Danish Maritime Authority \url{https://web.ais.dk/aisdata/aisdk-2024-11-20.zip}}, per-trip estimation of CO2 emissions~\footnote{Thetis-MRV data containing CO2 emissions information \url{https://mrv.emsa.europa.eu/#public/emission-report}}, as well as with the location of the different danish ports of which a list can be found at ~\footnote{\url{https://www.searates.com/maritime/denmark}}. The complete data is composed of the movement of 300 ships in the Baltic and North Sea around the country of Denmark during one day. %The accuracy of the frameworks is evaluated on subsets of ships of different sizes \mahmoud{what does this mean?}. 

%\gaspard{Here is the description of how the data was changed}
Due to computational constraints posed by Large Language Models, our approach involves pre-processing raw AIS data to minimize its size. We accomplished this by resampling the data at a 5-minute frequency, retaining only the most essential columns (mmsi, timestamp, latitude, longitude, and sog), and converting the timestamp to an HH:MM format, 
which limits the dataset to a single day's worth of data. By doing so, we further reduced the overall dataset size. Additionally, we merged columns containing static information such as draught or length with the CO2 emission dataset to create a single, static dataset for ship metadata.

We propose a classification of queries into four categories as described hereunder. Further, we propose a total of 27 queries divided into these categories. 

\subsection{Attribute Queries}
The questions/queries in this category are designed to retrieve or calculate specific attributes related to ships. These queries range from simple attribute retrievals to more complex reasoning and computation. Each query utilizes either direct lookups from enriched data tables or applies basic arithmetic calculations to derive new information. This set of attribute queries demonstrates the practical application of data manipulation and analysis techniques in understanding key characteristics of maritime transport vehicles, focusing on both individual ship details and aggregate data insights.

\begin{enumerate}[label=$\mathcal{Q}_\arabic*$:, leftmargin=*]
    \item What is the name of ship $[\mathit{MMSI}]$? This question asks for the name of a ship using its MMSI number, which is a unique identifier. You just look up this number in the AIS data table to find the ship's name.
    
    \item What is the IMO number for ship $[\mathit{MMSI}]$? This query requires finding the International Maritime Organization (IMO) number for a ship by using its MMSI number. The difference with the previous is in the datatype of the result, text v.s. long number, as perhaps this might impact the LLM performance.  

    \item What is the annual CO2 emission of ship $[\mathit{MMSI}]$? This is a straightforward lookup in the AIS data table, since we pre-process and enrich with the CO2 emission attribute.

    \item How many ships are in the dataset? This query groups all entries in the data table and counts how many unique ships there are.
    
    \item Assuming that the rectangle volume of a ship is a good approximation of its cargo capacity, which ship has the largest cargo capacity? This is a slightly more complex query, as it requires calculating the cargo capacity of each ship based on its dimensions (length, breadth, and height to approximate a rectangle's volume) and then determining which ship has the largest capacity.
        
    \item What is the CO2 emission of ship $[\mathit{MMSI}]$ per nautical mile ? This query requires simple inference to calculate the CO2 emission per nautical mile by linear referencing the annual CO2 emission.
    
    \item What is the volume of displaced water of ship $[\mathit{MMSI}]$? This query also requires simple inference to understand that the volume of water displaced by the ship is equivalent to the ship's volume under the waterline, and to perform this calculation. 
\end{enumerate}

%Two examples such query are $\mathcal{Q}_3$, \textit{``What is the name of ship $[MMSI=211870240]$?"} and $\mathcal{Q}_7$, \textit{``Assuming that the rectangle volume of a ship is a good approximation of its cargo capacity, which ship has the largest cargo capacity?"}.

\subsection{Individual Trajectory Queries}
These queries will evaluate the capacities of the LLM to perform aggregation operations and reasoning based on the trajectories of individual vessels. These will therefore require to analyse full trajectories and the evolution of the dynamic attributes over time.

    \begin{enumerate}[label=$\mathcal{Q}_{\arabic*}$:, resume, leftmargin=*]
      \item What is the maximum recorded speed of ship $[\mathit{MMSI}]$? reflecting on the vessel's operational performance and potential under specific conditions.
      
      \item Ship $[\mathit{MMSI}]$ is a ferry doing a round trip between two ports. It does it multiple times a day. What is the average single trip duration in minutes? Here, the focus is on analyzing the ferry trajectory, and understanding its commute pattern between the two ports. 
      
      \item What is the total number of kilometers traveled by ship $[\mathit{MMSI}]$ for the day? This query requires spatial computing of travel distance.
      
      \item What is the average speed of ship $[\mathit{MMSI}]$ in kilometers per hour based on all moments where the ship was navigating ($SOG > 0.1$) ?
      This query involves understanding the relation between the ship position and its $SOG$.
      
      \item What is the last known location of ship $[\mathit{MMSI}]$?
      This query involves reasoning about the temporal ordering of vessel location updates.  
      
      \item What is the average waiting time for ship $[\mathit{MMSI}]$ at anchorage? This query involves complex analysis to distinguish and segment the vessel trajectory into move and anchorage parts. This analysis is a topic for complete research papers, e.g.,~\cite{wu2022semantic} 
      
      \item For ferry $[\mathit{MMSI}]$, how many round trips did it complete?
      
    \end{enumerate}

%Two examples such query are provided by $\mathcal{Q}_8$, \textit{``What is the maximum recorded speed of ship $[MMSI=219028133]$?"} and $\mathcal{Q}_{13}$, \textit{``What is the average waiting time for ship $[MMSI=211190000]$ at anchorage?"}.

\subsection{Interaction Queries}
These queries will evaluate the capacities of LLM to analyze trajectory interaction with other vessels and port infrastructure.  
%These are however limited to the trajectory and other dynamic attributes. They do not require to combine multiple data sources such as for the fusion queries (see hereunder).
%One example focusing on ships' interactions is query $\mathcal{Q}_{16}$, \textit{``Identify clusters of ships traveling together."}. 
%Another example of queries of this category requiring complex reasoning from the LLM is $\mathcal{Q}_{17}$, \textit{``Which ships are ferries based on their navigation pattern?"}.

    \begin{enumerate}[label=$\mathcal{Q}_{\arabic*}$:, resume, leftmargin=*]
      \item Which other ships did ship $[\mathit{MMSI}]$ get close to $(< 500m)$ during its trip? This query involves distance calculation between pairs of spatiotemporal trajectories.
      
      \item Identify clusters of ships traveling together. This query is a sophisticated spatiotemporal data mining task that aims to find groups of ships that move in close proximity over a certain period. It involves analyzing the collective movement patterns of multiple ships to detect clusters or convoys. To answer this query, we typically use efficient data structures and algorithms that can handle the complexity of dynamic spatial relationships, potentially incorporating machine learning techniques to classify and predict clustering behavior. 
      
      \item Which ships are ferry (always doing the same roundtrip) based on their navigation pattern? This query requires a complex analysis of navigation patterns to determine if a ship behaves like a ferry, characterized by repeated trips between ports. It involves examining the regularity and consistency of a ship's journey cycles over time, checking if it follows a consistent route that begins and ends at the same locations, typically within a predictable timetable.
    \end{enumerate}

\subsection{Data-fusion Queries}
These queries focus on data fusion, where multiple datasets—such as ship trajectories, port locations, and fuel consumption records—are integrated to answer nuanced questions about maritime operations.
%One example of such query is $\mathcal{Q}_{24}$, \textit{``What is the CO2 emission from $[PORT_1=Aarhus]$ to $[PORT_2=Skagen]$ for ship $[MMSI=211190000]$?"}. 

    \begin{enumerate}[label=$\mathcal{Q}_{\arabic*}$:, resume, leftmargin=*]
      \item How long did ship $[\mathit{MMSI}]$ stay at sea during the day (was not close or in a port)? This query necessitates fusing ship location data with port data to determine periods when the ship was not docked or near any port. 
      
      \item Which ship consumed most fuel between 15:00 and 16:00 on 2024-11-20? Here, we integrate spatiotemporal trajectory data with fuel consumption logs during the specified hour to identify the ship with the highest fuel usage.
      
      \item What is the average trip length from port Skagen to port Aarhus? Calculating this involves merging spatiotemporal trajectory data of ships with port location data to assess the distances traveled between these two specific ports.
      
      \item Hypothetically, could ship $[\mathit{MMSI}]$ navigate through the Suez canal? To tackle this query, we analyze the ship's current trajectory and heading data, inferring its potential to enter the Suez Canal, despite not having explicit canal location data within our dataset.
       
      \item Based on the travel distance of each ship for the day, and leveraging their respective fuel consumption, which ship polluted the most? This requires a comprehensive fusion of data regarding each ship's travel distance (from trajectory data) and their respective fuel consumption data to evaluate environmental impact.
      
      \item Identify the ship with the best fuel efficiency (CO2 per nautical mile). This query combines CO2 emission data with the distance each ship has traveled, calculated from spatiotemporal data, to find the most fuel-efficient vessel.
      
      \item What is the CO2 emission for one trip from Aarhus to Skagen for ship $[\mathit{MMSI}]$? This query requires analyzing the respective vessel trajectory in fusion with the ports data and the CO2 emission data.
      
      \item Which is the most visited port for the day in terms of number of different boats that visited it?  To determine this, we aggregate and compare the number of distinct ships that docked at each port, requiring a fusion of vessel trajectory data with ports and inducing visitation records.
      
      \item Were any ships in the dataset at risk of collision? 
      This critical safety query assesses collision risks by analyzing proximity events between ships, requiring a synthesis of multiple ships' trajectory data.
            
      \item Which ship did not go to sea today? This query examines ships' proximity to port locations throughout the day using their spatiotemporal data to identify those that remained docked.
    \end{enumerate}

\section{Methods}
In this study, we use the methods of Natural Language Interfaces to Databases (NLIDB) and Zero-shot data analysis (ZSA), as previously introduced. These methods are implemented by posing questions directly to a Large Language Model (LLM) using a method known as contextualization or prompting. This process involves formulating a query in natural language and supplementing it with relevant contextual information to guide the model's response.

\noindent\textbf{Contextualization:} In the realm of LLMs, contextualization, or prompting, involves providing the model with specific background information or a setup that leads into the question. This \emph{prompt} primes the model to understand the query's context and generate a relevant answer. For example, when querying about maritime activities, the prompt may include not only the question but also supplementary data like vessel coordinates, CO2 emissions, and information about ports.

\noindent\textbf{Prompt engineering for different methods:} The nature of the prompt can vary significantly between NLIDB and Zero-shot methods:
\begin{itemize}[leftmargin=*]
    \item NLIDB: requires more structured and specific information. For instance, if querying about a ship's location, the prompt need to include precise details about the attributes of location within the dataset.
    \item ZSA: relies on the model's pre-trained knowledge to infer answers from general queries without additional context. Here, the prompt might simply involve the raw data and a straightforward question about ship movements or CO2 emissions without further details.
\end{itemize}
    
\noindent\textbf{Challenges with spatiotemporal data:} handling spatiotemporal data, like ship coordinates, introduces specific challenges due to its voluminous nature. For instance, representing a single 2D coordinate might take about 10 characters or tokens in a model’s context, which can quickly consume the model's maximum context window.

\noindent\textbf{Why context window size matters:} because this is the maximum length of input the LLM can handle at one time. If the input exceeds the context window size, the model starts trimming from the start of the input. This gives the impression that the model has forgotten part of the input. Context window size is therefore crucial when dealing with big data. A larger window allows for more data to be considered in one go, enhancing the model's ability to make informed predictions or analyses. This is particularly important in our case, where the data involved is both complex and voluminous.

\noindent\textbf{Choosing the right LLM:} given the necessity for a long context window to handle our data, we selected Gemini, known for having the largest context window among available LLMs. This choice is strategic to ensure that the richness of spatiotemporal maritime data can be fully leveraged, allowing the model to access and utilize a more comprehensive data snapshot during its processing.
While there exist specialized LLMs for specific scientific domains (for instance GeoGalactica \cite{lin2023geogalactica} or K2 \cite{deng2024k2}), to the best of the authors' knowledge, no specialized LLM exists for mobility (or geospatial) data. 

In particular, we use Gemini-1.5-flash models as they provide a context window of up to 2 Million tokens. For a clearer understanding of how Gemini stands in comparison to other LLMs in terms of context window sizes, refer to Table~\ref{table:llms}, which lists common LLMs alongside their respective context window sizes, illustrating why Gemini is particularly suited for our analytical needs.

% \begin{table}[h]
% \centering
% \caption{Comparison of Context Window Sizes in Different Large Language Models}
% \label{table:llms}
% \begin{tabular}{|m{3.5cm}|m{2.5cm}|m{1.5cm}|}
% \hline
% \textbf{Model Name} & \textbf{Context Window Size} & \textbf{Release Year} \\ \hline
% OpenAI's GPT-3\footnote{\url{https://openai.com/research/gpt-3}} & 4,096 tokens & 2020 \\ \hline
% Google's T5-Large\footnote{\url{https://arxiv.org/abs/1910.10683 }} & 512 tokens & 2020 \\ \hline
% AI21 Labs' Jurassic-1\footnote{\url{https://www.ai21.com/research/jurassic-1}} & 8,192 tokens & 2021 \\ \hline
% OpenAI's GPT-4 & Estimated 12,288 tokens & Future \\ \hline
% Gemini-1.5-flash\footnote{\url{https://gemini.com/models/gemini-1.5-flash}} & 2,000,000 tokens & 2023 \\ \hline
% \end{tabular}
% \end{table}

\begin{table}[h]
\centering
\begin{tabular}{l l l}
%\hline
\textbf{Model Name} & \makecell[bl]{\textbf{Context} \\ \textbf{Window Size}} & \textbf{Release Year} \\ \hline
OpenAI's GPT-3\footnote{\url{https://openai.com/research/gpt-3}} & 4,096 tokens & 2020 \\ 
Google's T5-Large\footnote{\url{https://arxiv.org/abs/1910.10683 }} & 512 tokens & 2020 \\ 
AI21 Labs' Jurassic-1\footnote{\url{https://www.ai21.com/research/jurassic-1}} & 8,192 tokens & 2021 \\ 
OpenAI's GPT-4 & \makecell[tl]{Estimated \\ 12,288 tokens} & Future \\ 
Gemini-1.5-flash\footnote{\url{https://gemini.com/models/gemini-1.5-flash}} & 2,000,000 tokens & 2023 \\ 
\end{tabular}
\caption{Comparison of Context Window Sizes in Different Large Language Models.}
\label{table:llms}
\end{table}

%LLMs have emerged very recently as massively used tools. Many models have been developed by major actors of the industry and the state of the art models are evolving rapidly. 
%As such, this work is not aimed at providing a comparison of different models. 
%Instead, we focus on one state-of-the-art model and illustrate different frameworks leveraging its capacities.
%Since one coordinate can take up to 10 tokens, it was decided to pick a model with a large context window. 
% One of the main concerns when handling AIS data with an LLM directly is the size of the context.

%In particular, we focus on Gemini models as they provide a context window of up to 2 Million tokens, while for instance, OpenAI's best models (4o/o1) have only 128,000 tokens. Geminis offer many models, for now, we use gemini-1.5-flash, we will see about using pro or 2.0-flash.
%Queries and data are publicly available at \todo{}, offering the reader the possibility of analyzing the capabilities of different models.

%Furthermore, this study will not focus on the fine tuning of LLM models. 
\noindent\textbf{Temperature parameter in LLMs: }
controls the randomness of predictions/answers given by an LLM. It is a scalar value that affects the distribution from which the next word is sampled during text generation. Low Temperature (e.g., 0.0 or 0.1) makes the model's responses more deterministic and repetitive. At a temperature close to zero, the LLM is more likely to choose the most probable next word from its vocabulary, leading to more predictable and conservative outputs. High Temperature (e.g., 0.8 to 1.0) increases randomness, making the model's responses more diverse and less predictable. This can be useful for generating creative content or when you want the model to explore a wider array of linguistic possibilities.

The choice of temperature often depends on the specific application. For example, in a scenario where accuracy and precision are critical, such as generating code or formal reports, a lower temperature might be preferred. In contrast, for creative writing or brainstorming sessions, a higher temperature could stimulate more novel outputs. In our experiments, since we don't want to introduce bias, a temperature value of 0.5 was selected, and combined with \emph{self-consistency}.

\noindent\textbf{Self-consistency method: }
is a technique used to enhance the reliability of the outputs from an LLM~\cite{wang2022self}. It is particularly useful when dealing with high temperatures that introduce more variability in the model's responses. The LLM is queried multiple times with the exact same input prompt. This is done to generate a variety of different outputs, each potentially capturing different aspects or interpretations of the input. Once multiple responses are obtained, they are aggregated to form a final answer. The aggregation method depends on the nature of the query. For \emph{numerical responses}, one way to aggregate is to take the median of all responses. The median helps minimize the impact of outliers. For \emph{textual responses}, the most frequently occurring response might be chosen as the most likely or reliable answer.

By combining a chosen temperature setting with the self-consistency method, users can tailor the performance of an LLM to meet specific requirements of both reliability and creativity, enhancing the model’s applicability across a wide range of tasks and domains.

%Indeed, some specific parameters can be set on different models. For instance, most model accept a temperature parameter, influencing the randomness in the generated results.
%A temperature of zero meaning that the LLM are fully deterministic, while a higher value lead to more variability. 
%In our experiments, a value of 0.5 was selected.
% snother possibility would have been to set temperature to 0, leading to deterministic results but with higher probabilities of errors. 
%The technique of self-consistency \cite{wang2022self} is used to take advantage of this variability to improve the accuracy of the results while still providing stable answers.
%This technique consists of querying the LLM a predefined number of times with the same input, and to aggregate its outputs in a final result. 
%The aggregating method depends on the query. 
%It can consist of taking the median, if the expected number is a numerical value, or simply taking the most frequent answer.
% To counter the effect of this variability, we use the self-consistency technique. This allows us to set a temperature set to 0.5 to allow the model to try different approaches and still obtain stable enough results. 
% would have rendered this pointless as all answers would have been the same.

\begin{figure}
    \centering
    \includegraphics[width=1\linewidth]{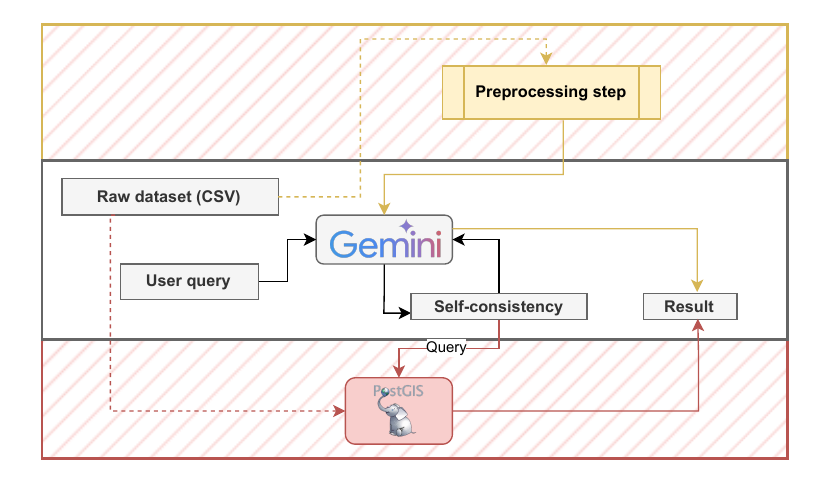}
    \caption{A schematic view of the different approaches. The $\mathit{ZSA}$ systems are represented by yellow and black rectangles. For each $\mathit{ZSA}$, only the preprocessing step differs. 
    The $\mathit{NLIDB}$ system is represented by the red and black boxes.}
    \label{fig:frameworks}
\end{figure}

Hereunder, four categories of LLM query methods are proposed. These are illustrated on Figure~\ref{fig:frameworks}.
One of them is NLIDB (using LLMs to issue queries to an existing GIS infrastructure), and the other three are variants of ZSA. % Three of them are $LAG$ frameworks 
(feeding plain text trajectories, feeding simplified plain text trajectories, feeding enriched semantic trajectories). 
The four methods are described in the following sections.

\subsection{$\mathit{NLIDB}$ Querying LLM with existing GIS infrastructure}
% This method investigates a similar approach as in \cite{li2023autonomous, zhang2024geogpt}. The data will be stored into a GIS system. The ``schema" of the data as well as access to this GIS system will be provided to the LLM as part of the prompt. 
The role of the LLM in this method consists in:
(1) the translation of the human language formulated query into the appropriate pipeline of operations (according to the GIS system); (2) the execution of the produced pipeline; and
(3) the interpretation of the produced results.
For this paper, data is stored on a PostgreSQL database with the PostGIS extension. The LLM is asked to produce SQL queries mapping to the 27 queries in this paper.

\subsection{$\mathit{ZSA}_1$: Querying LLM with plain AIS data}
In this method, AIS data is stored in three simple CSV files that will be passed to the LLM with the query in the prompt.
As part of the prompt engineering, the raw collected AIS data is split in two parts. 
The first part consists in the \emph{static attributes} of each trajectory which remain constant during the vessels' trips (for instance, the vessel's name and dimensions).
The second part consists in the time-varying movement data of the vessels, changing along the trips, such as the position and the speed over ground SOG.

This separation is performed in order to avoid feeding the LLM duplicate data, and therefore increase the quantity of data that could fit in its context.
Furthermore, a third CSV file containing the location of relevant ports is also provided to the LLM.
This method is intended to be a baseline for comparing the other methods. It helps investigating LLMs' capacities at understanding and processing raw geospatial data and operations.

\subsection{$\mathit{ZSA}_2$ Querying LLM with simplified trajectories}
A major concern in feeding spatiotemporal data to the LLMs is the limitation imposed by the size of the LLMs contexts, in contrast to the large nature of spatiotemporal data.
Basically, this limits the number of trajectories that can be fed to a LLM before it starts to forget the trajectories provided earlier.
For this reason, $\mathit{ZSA}_2$ uses trajectory compression to reduce the number of points in the trajectory before passing it to the LLM. This method thus investigates the capacities of a LLM to handle compressed trajectories.

The lossy compression of trajectories can facilitate LLMs to process a larger number of trajectories within their context. To achieve this, we employed the Top Down Time Ratio (TDTR) algorithm, which has been widely used for trajectory compression. 
Specifically, we utilized the Python library MovingPandas, which provides an implementation of TDTR through its -\textit{TopDownTimeRatioGeneralizer} class. 
The resulting compressed trajectories allow the LLM to handle a higher volume of data.

\subsection{$\mathit{ZSA}_3$ Querying LLM with semantic trajectories}
In this method, we investigate whether the performances of the LLMs can be improved by transforming the raw trajectories in a sequence of semantic events, closer to the natural language than raw coordinates.
We defined these events as the transition from one known geographical area, such as a port or a river, to another.  Then we used Algorithm \ref{alg:trajectory_to_events} to transform a given trajectory into a sequence of semantic events. This algorithm is similar to commonly used trajectory annotation, as in \cite{ liu2021semantics}.
It works as follows. For each trajectory point in a specific trajectory, the zone in which the point lies is computed. These zones correspond to the OSM regions. The zones are assessed from smallest to largest to ensure the most narrow and accurate possible description. 
If the zone of the current point is the same as the previous point, the distance traveled in this zone is updated.
Finally, in order to avoid constant oscillations for vessels navigating at the limit between two zones, a buffer distance is used.

\begin{comment}
    
\gaspard{Should we enumerate OSM sources?} 
\gaspard{CLOSEST: https://dl.acm.org/doi/10.1145/1951365.1951398 and https://infoscience.epfl.ch/server/api/core/bitstreams/982caf59-d965-4392-8fc1-2297c498704e/content}
\gaspard{https://link.springer.com/article/10.1007/s41060-024-00614-w}
\gaspard{more geneirc: https://link.springer.com/article/10.1007/s41651-021-00088-5}
\gaspard{https://www.tandfonline.com/doi/abs/10.1080/13658816.2020.1798966}

\begin{algorithm}
\caption{Trajectory to Semantic Events Conversion}
\label{alg:trajectory_to_events}
\begin{algorithmic}[1]
\REQUIRE List of geographic zones (bounding boxes)  ordered by size
\REQUIRE Trajectory
\REQUIRE Buffer threshold  
\ENSURE List of semantic events  

\FOR{each trajectory point}
    \FOR{each zone}
        \IF{closer than buffer}
            \STATE Break, current zone found
        \ENDIF
    \ENDFOR
    \IF{the zone is the same as last run}
        \STATE Accumulate distance traveled  
    \ELSE
        \STATE Record the previous zone, total distance, and duration  
        \STATE Update to the new zone  
        \STATE Reset cumulative distance  
    \ENDIF
\ENDFOR

\RETURN List of semantic events
\end{algorithmic}

\end{algorithm}

\end{comment}

\begin{algorithm}
\caption{Trajectory to Semantic Events Conversion}
\label{alg:trajectory_to_events}

\SetKwInOut{Input}{Require}
\SetKwInOut{Output}{Ensure}

\Input{List of geographic zones ordered by size}
\Input{Trajectory}
\Input{Buffer threshold}
\Output{List of semantic events}

\SetKwFor{For}{for}{do}{end} % Hide the 'end'
\SetKwIF{If}{ElseIf}{Else}{if}{then}{else if}{else}{end} % Hide the 'end'

\For{each trajectory point}{
    \For{each zone bounding box}{
        \If{closer than buffer}{
            Break, current zone found
        }
    }
    \If{the zone is the same as last run}{
        Accumulate distance traveled
    }
    \Else{
        Record the previous zone, total distance, and duration\\
        Update to the new zone\\
        Reset cumulative distance
    }
}
\Return{List of semantic events}
\end{algorithm}

\section{Ground-Truth and Evaluation}
For the 27 queries developed in Section~\ref{sec:queries}, we created a robust ground truth by employing a combination of methods. First, we used Python and SQL scripts to derive precise answers to the questions. Additionally, for queries necessitating subjective judgment, we consulted domain experts to ensure accuracy and relevance.

% We evaluated each method's performance by scoring its responses to the queries. For every query where the method's response matched the ground truth exactly, we awarded one point. If a response didn't fully match but was partially correct, it received half a point; incorrect answers received zero points.

Some queries in our set specifically required identification details, such as the MMSI (Maritime Mobile Service Identity) of one or several ships. For these, we tested the methods' ability to accurately respond to the same query for different MMSI values. The score for these queries was calculated based on the proportion of MMSI values for which the method provided the correct answer. 

\section{Empirical Results}

The average results of the four methods on all queries for different sizes of datasets is illustrated in Figure~\ref{fig:aggregated_model}. 
As expected, we can see that the $\mathit{NLIDB}$ model results remain stable with the size of the dataset, 
While all of $\mathit{ZSA}_1$, $\mathit{ZSA}_2$ and $\mathit{ZSA}_3$ performances quickly decrease with the size of the dataset.
It was expected that compression and transformation of the trajectories in semantic events would improve the capacities of the method (or at least reduce the degradation with larger datasets) with respect to feeding raw data. 
We can see however that this is not the case.
While the accuracies of all $\mathit{ZSA}$ models are pretty similar for a larger number of trajectories, $\mathit{ZSA}_1$ is significantly outperforming the two other $\mathit{ZSA}$ methods when analysing only a few trajectories.
\begin{figure}
\pgfplotstablegetrowsof{assets/Compressed.csv}
\edef\numberofrows{\pgfplotsretval}
\begin{tikzpicture}
    \begin{axis}[
    height=4cm,
    width=0.4\textwidth, scale only axis, % To fit subfigures
    xtick={5,10,25,50, 75, 100},
    ylabel={Score},
    xmin=-0.5,
    xlabel={Dataset Size},
    xmax=105,
    ymin=0,
    ymax=1.1,
    ]
        % First plot
        %\node[draw,fill=white,inner sep=3pt,above left=0.5em] at (60, 0.9) {Raw};
        \addplot[thick, mark options={fill=blue},   mark=square*, color=blue, discard if not={model}{Raw}] table[x index=0, y index=2, col sep=comma] {assets/all.csv};
        \addplot[thick, mark options={fill=red},    mark=o, color=red, discard if not={model}{Compressed}] table[x index=0, y index=2, col sep=comma] {assets/all.csv};
        \addplot[thick, mark options={fill=green},  mark=triangle*, color=green, discard if not={model}{Semantic}] table[x index=0, y index=2, col sep=comma] {assets/all.csv};
        \addplot[thick, mark options={fill=orange}, mark=x, color=orange, discard if not={model}{PostGIS}] table[x index=0, y index=2, col sep=comma] {assets/all.csv};

    \end{axis}
\end{tikzpicture}
    \caption{Accuracy of the frameworks. The frameworks $ZSA_1$ (raw), $ZSA_2$ (simplified), $ZSA_3$ (semantic) and $NLIDB$ (PostGIS) are represented in blue, red, green and orange respectively.}
    \label{fig:aggregated_model}
\end{figure}
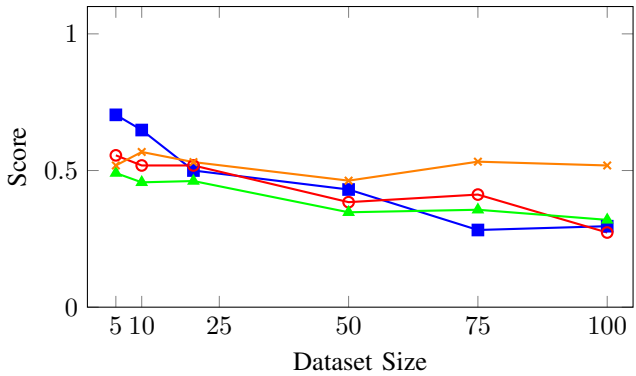

% Figure \ref{fig:accuracy_query_type} 
\begin{figure*}
  \center
\pgfplotstablegetrowsof{assets/Compressed.csv}
\edef\numberofrows{\pgfplotsretval}
\begin{tikzpicture}
    \begin{groupplot}[
        group style={
            group size=2 by 2, % One column, two rows
            vertical sep=0pt, % No vertical space
            horizontal sep=0pt, % No vertical space
        },
    height=4cm,
    width=0.4\textwidth, scale only axis, % To fit subfigures
    xtick={5,10,25,50, 75, 100},
    xmin=-0.5,
    xmax=105,
    ymin=0,
    ymax=1.1,
]
        % First plot
        \nextgroupplot[
        ylabel={Average Score},
        xticklabels={},
        ]
        \node[draw,fill=white,inner sep=3pt,above left=0.5em] at (60, 0.9) {Attribute Queries};
        \addplot[thick, mark options={fill=blue}, mark=square*, color=blue, discard if not={model}{Raw}] table[x index=1, y index=2, col sep=comma] {assets/attribute.csv};
        \addplot[thick, mark options={fill=red}, mark=o, color=red, discard if not={model}{Compressed}] table[x index=1, y index=2, col sep=comma] {assets/attribute.csv};
        \addplot[thick, mark options={fill=green}, mark=triangle*, color=green, discard if not={model}{Semantic}] table[x index=1, y index=2, col sep=comma] {assets/attribute.csv};
        \addplot[thick, mark options={fill=orange}, mark=x, color=orange, discard if not={model}{PostGIS}] table[x index=1, y index=2, col sep=comma] {assets/attribute.csv};

        % Second plot (shares x-axis)
        \nextgroupplot[ylabel={}, xticklabels={}, yticklabels={},
        ]
        \node[draw,fill=white,inner sep=3pt,above left=0.5em] at (85, 0.9) {Individual Trajectory Queries};
        \addplot[thick, mark options={fill=blue}, mark=square*, color=blue, discard if not={model}{Raw}] table[x index=1, y index=2, col sep=comma] {assets/individual-trajectory.csv};
        \addplot[thick, mark options={fill=red}, mark=o, color=red, discard if not={model}{Compressed}] table[x index=1, y index=2, col sep=comma] {assets/individual-trajectory.csv};
        \addplot[thick, mark options={fill=green}, mark=triangle*, color=green, discard if not={model}{Semantic}] table[x index=1, y index=2, col sep=comma] {assets/individual-trajectory.csv};
        \addplot[thick, mark options={fill=orange}, mark=x, color=orange, discard if not={model}{PostGIS}] table[x index=1, y index=2, col sep=comma] {assets/individual-trajectory.csv};

        % Third plot (shares x-axis)
        \nextgroupplot[, xticklabels={},
        ylabel={Average Score},
        xticklabels ={5, 10, 20, 50, 75, 100},
        xlabel={Dataset Size},
        ]
        \node[draw,fill=white,inner sep=3pt,above left=0.5em] at (75, 0.9) {Interactions Queries};
        \addplot[thick, mark options={fill=blue}, mark=square*, color=blue, discard if not={model}{Raw}] table[x index=1, y index=2, col sep=comma] {assets/ship-interaction.csv};
        \addplot[thick, mark options={fill=red}, mark=o, color=red, discard if not={model}{Compressed}] table[x index=1, y index=2, col sep=comma] {assets/ship-interaction.csv};
        \addplot[thick, mark options={fill=green}, mark=triangle*, color=green, discard if not={model}{Semantic}] table[x index=1, y index=2, col sep=comma] {assets/ship-interaction.csv};
        \addplot[thick, mark options={fill=orange}, mark=x, color=orange, discard if not={model}{PostGIS}] table[x index=1, y index=2, col sep=comma] {assets/ship-interaction.csv};

        % Forth plot (shares x-axis)
        \nextgroupplot[ylabel={}, yticklabels={},
        xticklabels ={5, 10, 20, 50, 75, 100},
        xlabel={Dataset Size},
        ]
        \node[draw,fill=white,inner sep=3pt,above left=0.5em] at (60, 0.9) {Fusion Queries};
        \addplot[thick, mark options={fill=blue}, mark=square*, color=blue, discard if not={model}{Raw}] table[x index=1, y index=2, col sep=comma] {assets/fusion.csv};
        \addplot[thick, mark options={fill=red}, mark=o, color=red, discard if not={model}{Compressed}] table[x index=1, y index=2, col sep=comma] {assets/fusion.csv};
        \addplot[thick, mark options={fill=green}, mark=triangle*, color=green, discard if not={model}{Semantic}] table[x index=1, y index=2, col sep=comma] {assets/fusion.csv};
        \addplot[thick, mark options={fill=orange}, mark=x, color=orange, discard if not={model}{PostGIS}] table[x index=1, y index=2, col sep=comma] {assets/fusion.csv};

    \end{groupplot}
\end{tikzpicture}
    \caption{Accuracy of the frameworks according to types of query.
            %The number of correct and incorrect answers are represented respectively in blue and red.}
            The scores of the frameworks $\mathit{ZSA}_1$ (Raw), $\mathit{ZSA}_2$ (Compressed), $\mathit{ZSA}_3$ (Semantic) and $\mathit{NLIDB}$ (PostGIS) are represented in blue, red, green and orange.}
    \label{fig:accuracy_query_type}
\end{figure*}
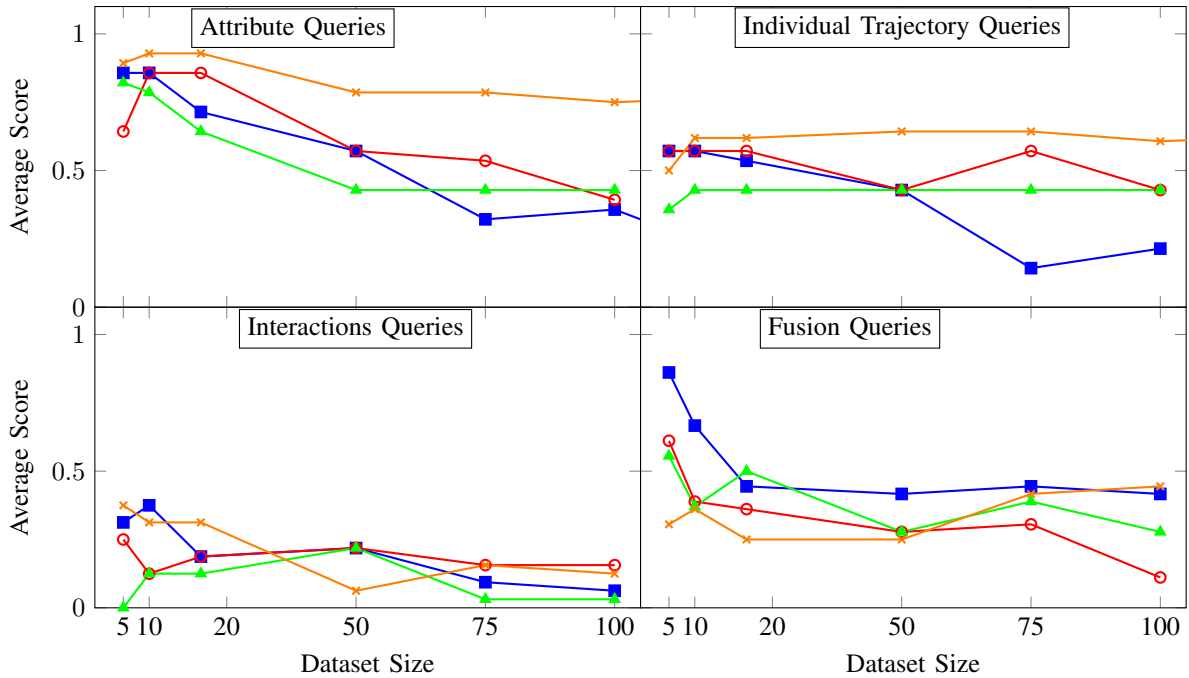
Figure \ref{fig:accuracy_query_type} details the results per the four classes of queries.
While the $\mathit{NLIDB}$ method performs globally well, it struggles answering the fusion queries.
We can also notice that none of the systems showcase a good accuracy for ship interaction related queries.
In particular, it should be noted that only $\mathcal{Q}_{16}$  has consistently been answered wrongly by all methods while on the opposite, $\mathcal{Q}_{12}$ is the only query that was always answered correctly by all methods.

\begin{figure}
    \centering
    % \begin{minipage}{0.49\textwidth}
        \centering
        \includegraphics[width=0.4\textwidth]{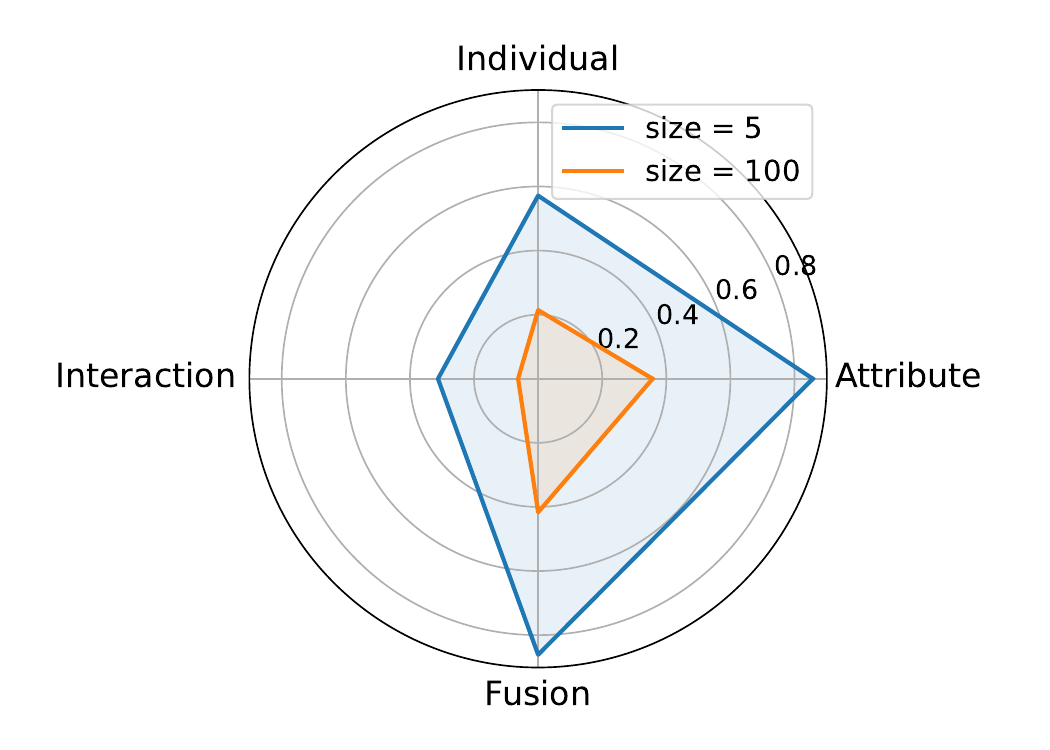}
    \caption{Performances of $\mathit{ZSA}_1$ for datasets of sizes 5 and 100.}
    \label{fig:radar_raw}
    % \end{minipage}
\end{figure}
    % \hfill
\begin{figure}
    % \begin{minipage}{0.49\textwidth}
        \centering
        \includegraphics[width=0.40\textwidth]{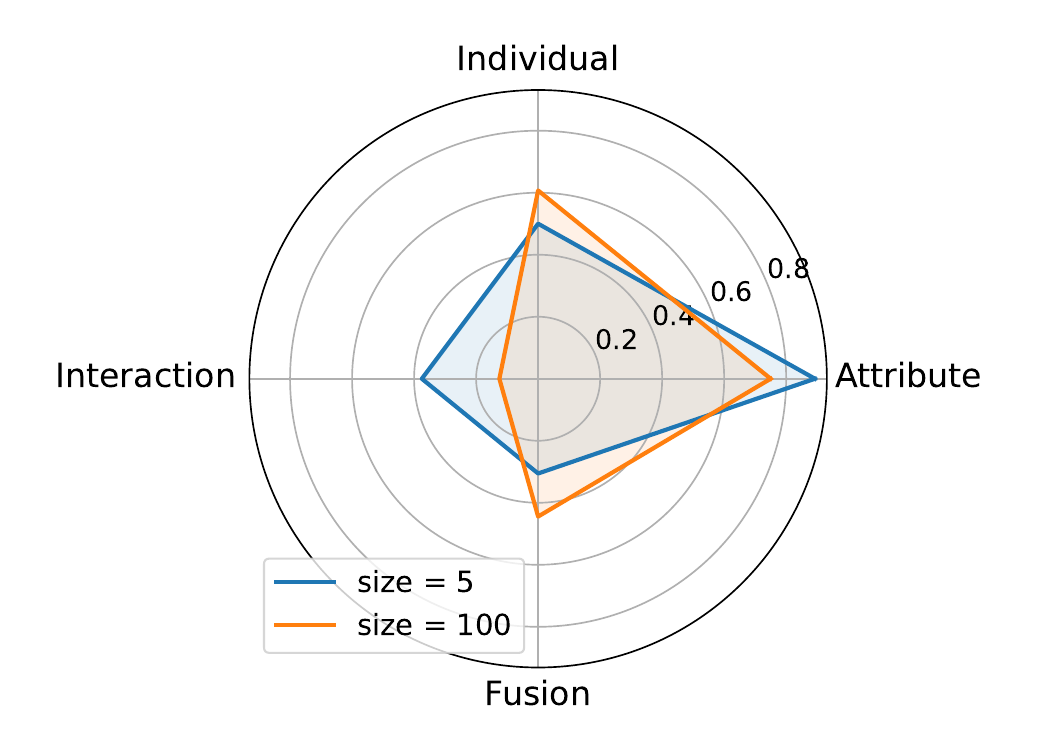}
    % \end{minipage}
    \caption{Performances of $\mathit{NLIDB}$ for datasets of sizes 5 and 100.}
    \label{fig:radar_postgis}
\end{figure}

The behaviour of $\mathit{ZSA}_1$ and $\mathit{NLIDB}$ are illustrated in Figures~\ref{fig:radar_raw} and \ref{fig:radar_postgis}. This figure represents two radar charts illustrating the scores of $\mathit{ZSA}_1$ and $\mathit{NLIDB}$ methods for the smallest and largest tested datasets. 
This figure furthers illustrates the change in performances of $\mathit{ZSA}_1$ according to the size of the dataset while $\mathit{NLIDB}$'s performances remain stable.

While ZSA methods performed worst in general for larger dataset sizes, they performed better than the NLIDB one for reduced dataset size. In our experiments, these methods have proven to be able to perform complex reasoning on a few trajectories composed of raw positional data, such as counting the round trips of a ferry.

This is especially true for $\mathit{ZSA}_1$, which provides by far, the best accuracy of all methods. $\mathit{ZSA}_1$ method directly ingests the raw data, without any other form of pre-treatment. This could imply that LLMs, or at least gemini-2.0-flash, are more than capable of understanding raw data, and do not need semantic sugar to help them do so.

That being said, all methods also show some limitations in the complexity of the queries they can handle, in particular none of the methods were able to identify clusters of ships traveling together. 

In conclusion, we have experienced that LLMs are able, while facing some serious limitations, to answer complex spatiotemporal queries, and in fact, to act as GIS. The fact that they can simply ingest the data in their context and start answering questions is what makes this ZSA approach so powerful. LLMs can simply take any human-readable format and start answering queries about it, all that is required is some powerful GPUs.  

Moreover, with the constant increase in context size and advancement in reasoning, LLMs will probably grow better at such a task, hence reducing further the gap between GIS operated by knowledgeable workers and LLM as GIS. This could allow more people to interact with highly complex data, without much technical skills.

Finally, while this research paves the way for highly interesting future developments, we must underscore the risk that LLMs can pose when blindly trusted, hence any public-facing development should always bear that limitation in mind.

\section{conclusion}
%In this work, we investigated the capacities of different systems based on LLMs to analyse and answer queries based on a provided dataset. 
%We showed that LLMs can perform complex reasoning on a few trajectories composed of raw positional data, such as, counting the round trips of a ferry.
%However, the performances of the frameworks based on the direct feeding of trajectory data to the LLM quickly degrade with the size of the dataset.
%We showed that compressing or transforming trajectories into semantic representation did not improve the accuracy of the framework as expected, but rather deteriorated them.
%As expected, the framework consisting in the generation of queries used to query a classical database system.
%All frameworks also show some limitations in the complexity of the queries they can handle, in particular none of the frameworks were able to identify clusters of ships travelling together. 

This study evaluated the performance of Natural Language Interface to Databases (NLIDB) and Zero-Shot Analysis (ZSA) methods using Large Language Models (LLMs) for analyzing Automatic Identification System (AIS) data across varying dataset sizes. The NLIDB method demonstrates consistent performance irrespective of dataset size, showcasing its stability. In contrast, the performance of the ZSA methods generally declines as the size of the dataset increases, indicating a challenge in handling larger volumes of data due to the context size. While it was hypothesized that transforming data into semantic events might improve ZSA performance, this did not significantly prevent performance degradation compared to using raw data.

The study also highlights that all methods struggle with complex queries, particularly those involving ship interactions, with none able to identify clusters of ships traveling together. This paper underscored the potential of LLMs to function as a Geographic Information System (GIS), capable of processing spatiotemporal queries directly from raw data formats. However, it also emphasized the limitations of current models in dealing with complex data queries and the necessity for cautious deployment given the risks of over-reliance on automated systems.

\begin{comment}
\mahmoud{no need for future work section}
\section{Future works}

Paths for future work:

1. Fine-tuned models 
2. Foundation model for mobility?
3. Leveraging multi-modal models to answer queries by visualizing the data
4. Agent approach where the model could be aware of its own limitation and query the data by part.
5. Linked to previous point: merge comprehension of raw data with GIS capabilities. To the heavy lifting with GIS, but still perform computation with own reasoning capabilities.
6. COT models ? 

\end{comment}

\bibliographystyle{IEEEtran}
\bibliography{bibli}

%\appendix  
%\section{Queries}
%\include{queries}

\end{document}